\documentclass{article}

\PassOptionsToPackage{numbers, compress}{natbib}

 \usepackage[preprint, nonatbib]{neurips_2020}

\usepackage[utf8]{inputenc} 
\usepackage[T1]{fontenc}    
\usepackage{hyperref}       
\usepackage{url}            
\usepackage{booktabs}       
\usepackage{amsfonts}       
\usepackage{nicefrac}       
\usepackage{microtype}      
\usepackage{amsmath,amssymb} 
\usepackage{graphicx}
\usepackage{multirow}
\usepackage{natbib}
\usepackage{subcaption}
\usepackage{multicol}
\usepackage{xcolor}
\usepackage[title]{appendix} 

\title{Visual Concept Reasoning Networks}

\author{
\textbf{Taesup Kim}$^{1}$\thanks{work also done while working at Kakao Brain.  Correspondence to {\tt taesup.kim@umontreal.ca}} , \textbf{Sungwoong Kim}$^{2}$, \textbf{Yoshua Bengio}$^{1}$
\\
$^1$Mila, Université de Montréal,  $^2$Kakao Brain
}
\begin{document}

\maketitle

\begin{abstract}
A split-transform-merge strategy has been broadly used as an architectural constraint in convolutional neural networks for visual recognition tasks.
It approximates sparsely connected networks by explicitly defining multiple branches to simultaneously learn representations with different visual concepts or properties.
Dependencies or interactions between these representations are typically defined by dense and local operations, however, without any adaptiveness or high-level reasoning.
In this work, we propose to exploit this strategy and combine it with our \textit{Visual Concept Reasoning Networks}~(VCRNet) to enable reasoning between high-level visual concepts.
We associate each branch with a visual concept and derive a compact concept state by selecting a few local descriptors through an attention module.
These concept states are then updated by graph-based interaction and used to adaptively modulate the local descriptors.
We describe our proposed model by split-transform-\textit{attend-interact-modulate}-merge stages, which are implemented by opting for a highly modularized architecture.
Extensive experiments on visual recognition tasks such as image classification, semantic segmentation, object detection, scene recognition, and action recognition show that our proposed model, VCRNet, consistently improves the performance by increasing the number of parameters by less than $1\%$.
\end{abstract}

\section{Introduction}
Convolutional neural networks have shown notable success in visual recognition tasks by learning hierarchical representations.
The main properties of convolutional operations, which are local connectivity and weight sharing, are the key factors that make it more efficient than fully-connected networks for processing images.
The local connectivity particularly comes up with a fundamental concept, receptive field, that defines how far the local descriptor can capture the context in the input image.
In principle, the receptive field can be expanded by stacking multiple convolutional layers or increasing the kernel size of them.
However, it is known that the effective receptive field only covers a fraction of the theoretical size of it~\cite{erf}.
This eventually restricts convolutional neural networks to capture the global context based on long-range dependencies.
On the other hand, most of convolutional neural networks are characterized by dense and local operations that take the advantage of the weight sharing property. 
It hence typically lacks internal mechanism for high-level reasoning based on abstract semantic concepts such those humans manipulate with natural language and inspired by modern theories of consciousness~\cite{Conscious}.
It is related to system 2 cognitive abilities, which include things like reasoning, planning, and imagination, that are assumed to capture the global context from interactions between a few abstract factors and accordingly give feedback to the local descriptor for decision-making.

There have been approaches to enhance capturing long-range dependencies such as non-local networks~\cite{NLN}.
The main concept of it, which is related to self-attention~\cite{NIPS2017_7181}, is to compute a local descriptor by adaptively aggregating other descriptors from all positions, regardless of relative spatial distance.
In this setting, the image feature map is plugged into a fully-connected graph neural network, where all local positions are fully connected to all others. 
It is able to capture long-range dependencies and extract the global context, but it still works with dense operations and lacks high-level reasoning. 
Both LatentGNN~\cite{latentgnn} and GloRe~\cite{glore} alleviate these issues by introducing compact graph neural networks with some latent nodes designed to aggregate local descriptors.

In this work, we propose \textit{Visual Concept Reasoning Networks} (VCRNet) to enable reasoning between high-level visual concepts.
We exploit a modularized multi-branch architecture that follows a split-transform-merge strategy~\cite{resnext}.
While it explicitly has multiple branches to simultaneously learn multiple visual concepts or properties, it only considers the dependencies or interactions between them by using dense and local operations.
We extend the architecture by split-transform-\textit{attend-interact-modulate}-merge stages, and this allows to capture the global context by reasoning with sparse interactions between high-level visual concepts from different branches.

The main contributions of the paper are:
\begin{itemize}
    \item We propose Visual Concept Reasoning Networks (VCRNet) that efficiently capture the global context by reasoning over high-level visual concepts.
    \item We compactly implement our propose model by exploiting a modularized multi-branch architecture composed of split-transform-attend-interact-modulate-merge stages. 
    \item We showcase our proposed model improves the performance more than other models by increasing the number of parameters by less than $1\%$ on multiple visual recognition tasks.
\end{itemize}

\section{Related Works}
Multi-branch architectures are carefully designed with multiple branches characterized by different dense operations, and split-transform-merge stages are used as the building blocks.
The Inception models~\cite{inception} are one of the successful multi-branch architectures that define branches with different scales to handle multiple scales.
ResNeXt~\cite{resnext} is another version of ResNet\cite{resnet} having multiple branches with the same topology in residual blocks, and it is efficiently implemented by grouped convolutions.
In this work, we utilize this residual block and associate each branch of it with a visual concept.

There have been several works to adaptively modulate the feature maps based on the external context or the global context of input data.
Squeeze-and-Excitation networks (SE)~\cite{SENET} use gating mechanism to do channel-wise re-scaling in accordance with the channel dependencies based on the global context.
Gather-Excite networks (GE)~\cite{GENET} further re-scale locally that it is able to finely redistribute the global context to the local descriptors.
Convolutional block attention module (CBAM)~\cite{CBAM} independently and sequentially has channel-wise and spatial-wise gating networks to modulate the feature maps.
All these approaches extract the global context by using the global average pooling that equally attends all local positions.
Dynamic layer normalization (DLN)~\cite{DLN} and Feature-wise Linear Modulation (FiLM)~\cite{FILM} present a method of feature modulation on normalization layers by conditioning on the global context and the external context, respectively. 

Content-based soft-attention mechanisms~\cite{attention} have been broadly used on neural networks to operate on a set of interchangeable objects and aggregate it. 
Particularly, Transformer models~\cite{NIPS2017_7181} have shown impressive results by using multi-head self-attention modules to improve the ability of capturing long-range dependencies. 
Non-local networks (NL)~\cite{NLN} use this framework in pixel-level self-attention blocks to implement non-local operations.
Global-context networks (GC)~\cite{GCNET} simplify the non-local networks by replacing the pixel-level self-attention with a attention module having a single fixed query that is globally shared and learned.
Attention-augmented convolutional networks~\cite{AA} similarly augment convolutional operators with self-attention modules as the non-local networks, but concatenate feature maps from convolution path and self-attention path.
LatentGNN~\cite{latentgnn} and Global reasoning module (GloRe)~\cite{glore} differently simplifies the non-local networks that they first map local descriptors into latent nodes, where the number of nodes is relatively smaller than the number of local positions, and capture the long-range dependencies from interactions between the latent nodes. 
Our proposed model is similar to these two models, but we take the advantage of the multi-branch architecture and the attention mechanism to efficiently extract a set of distinct visual concept states from the input data.

\begin{figure}[t]
\centering
\includegraphics[width=0.8\textwidth]{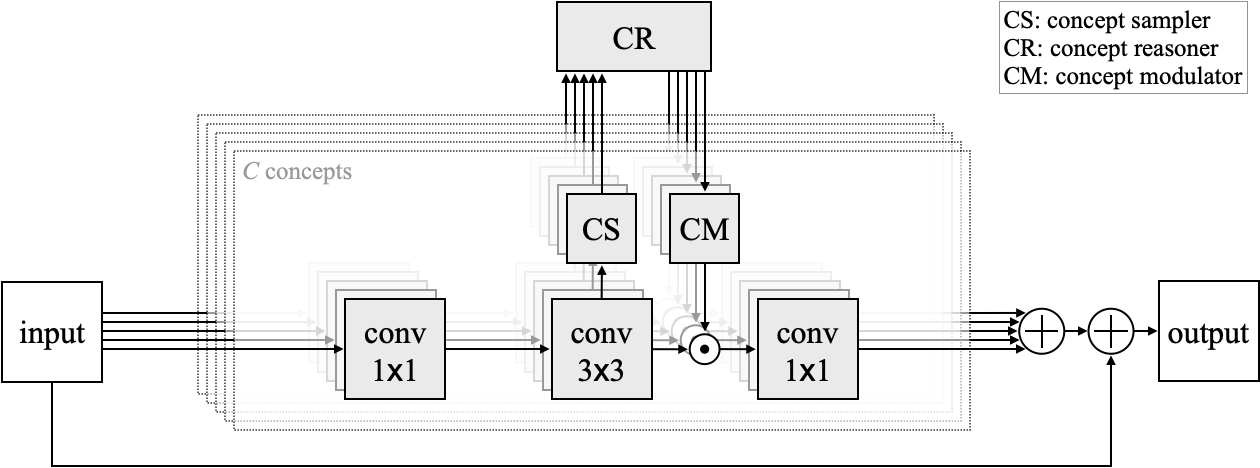}
\caption{A residual block with visual concept reasoning modules.}
\label{fig:vcr}
\end{figure}

\section{Methods}
In this section, we introduce our proposed model, Visual Concept Reasoning Network (VCRNet), and describe the overall architecture and its components in detail.
The proposed model is designed to reason over high-level visual concepts and accordingly modulate feature maps based on its result.
In the following, we assume the input data $X \in \mathbb{R}^{HW \times d}$ is a 2D tensor as an image feature map, where $H, W$, and $d$ refer to the height, width, and feature size of $X$, respectively.
Moreover, for simplicity, we denote all modules by a function $F_{\text{func}}(\cdot; \theta)$, where $\theta$ is a learnable parameter and the subscript func briefly explains the functionality of it.

\subsection{Modularized Multi-Branch Residual Block}
Residual blocks are composed of a skip connection and multiple convolutional layers~\cite{resnet}. 
We especially take advantage of using a residual block of ResNeXt~\cite{resnext} that operates by grouped convolutions. 
This block is explicable by a \textit{split-transform-merge} strategy and a highly modularized multi-branch architecture.
It has an additional dimension "cardinality" to define the number of branches used in the block.
The branches are defined by separate networks, which are based on the same topology and implemented by grouped convolutions, processing non-overlapping low-dimensional feature maps.
In this work, we use this block by regarding each branch as a network learning representation of a specific \textit{visual concept} and, therefore, refer to the cardinality as the number of visuals concepts $C$. 

The split-transform-merge strategy can be described by visual concept processing as the following.
Each concept $c$ has a compact concept-wise feature map $Z_c \in \mathbb{R}^{HW \times p}$, where $p$ is a lot smaller than $d$.
It is initially extracted from the input data $X$ by \textit{splitting} it into a low-dimensional feature map $\tilde{X}_c  \in \mathbb{R}^{HW \times p}$ with a $1 \times 1$ convolution $F_{\text{split}}(X; \theta^{\text{split}}_c)$.
Afterward, it is followed by a \textit{concept-wise transformation} based on a $3 \times 3$ convolution $F_{\text{trans}}(\tilde{X}_c;\theta^{\text{trans}}_c)$ while keeping the feature size compact.
The extracted concept-wise feature maps $\{Z_c\}_{c=1}^{C}$ are then projected back into the input space to be \textit{merged} as $Y=X + \sum_{c=1}^{C}{F_{\text{merge}}(Z_c; \theta^{\text{merge}}_c)}$. 
This overall multi-branch procedure interestingly can be highly modularized and parallelized by grouped convolutions.
However, it lacks the ability of reasoning over the high-level visual concepts that captures both local and global contexts.
We propose to extend this approach by introducing additional modules to enable visual concept reasoning.
Our proposed model is based on a new strategy with split-transform-\textit{attend-interact-modulate}-merge stages.
The new stages completely work into the residual block with the following modules: (a) concept sampler, (b) concept reasoner, and (c) concept modulator.
The overall architecture is depitced in Figure~\ref{fig:vcr} showing how it is highly modularized by sharing the topology between different concepts.
We refer to networks having residual blocks with these modules, as Visual Concept Reasonining Networks (VCRNet).

\begin{figure}[t]
\centering
\includegraphics[width=0.85\textwidth]{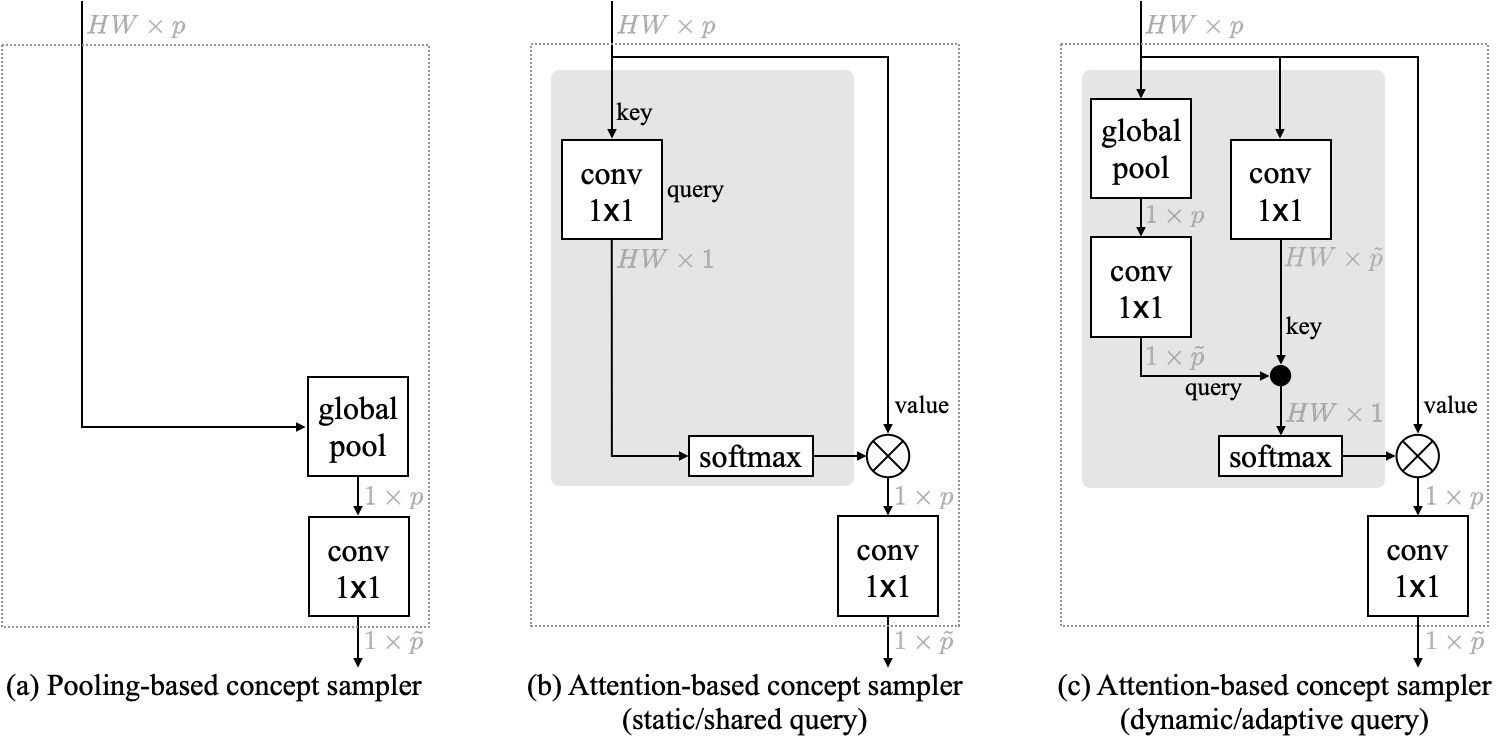}
\caption{Concept samplers with different approaches ($\otimes$ is a weighted-sum operation).}
\label{fig:cs}
\end{figure}

\subsection{Concept Sampler}\label{sec:cs}
The concept-wise feature maps $\{Z_c\}_{c=1}^{C}$ are composed of all possible pixel-level local descriptors, which contain spatially local feature information, as sets of vectors.
To do efficient reasoning over the visual concepts, it first requires a set of abstract feature vectors representing the visual concepts.
Therefore, a form of aggregation mechanism is necessary to derive a set of visual concept states, where each state is a vector, from the concept-wise feature maps.
We implement this by presenting a \textit{concept sampler}~(CS) module.
Each concept $c$ has a separate concept sampler $F_{\text{CS}}(Z_c; \theta^{\text{CS}}_c)$ that aggregates the set of local descriptors in $Z_c$ and converts it into a concept state $h_c \in \mathbb{R}^{1 \times \tilde{p}}$, where we set $\tilde{p}=\text{min}(p/4, 4)$.
We introduce two types of concept samplers that are based on pooling and attention operations, respectively.

\textbf{Pooling-based sampler }
Global average pooling is one of the simplest ways to extract the global context from a feature map without explicitly capturing long-range dependencies. 
It equally and densely attends all local positions to aggregate the local descriptors.
Our pooling-based sampler adopts this operation to compute the concept state $h_c$ as shown in Figure~\ref{fig:cs}.a, and it is formulated as:
\begin{equation}
h_c =F_{\text{CS}}(Z_c; \theta^{\text{CS}}_c) = F_{\text{GAP}}(Z_c) W^{\text{v}}_c = \left(\frac{1}{HW}\sum_{i=1}^{H}\sum_{j=1}^{W}Z_{c}[i, j]\right)W^{\text{v}}_c,    
\end{equation}
where $Z_c[i,j] \in \mathbb{R}^{1 \times p}$ is a local descriptor at position $(i, j)$, and $W^{\text{v}}_c \in \mathbb{R}^{p \times \tilde{p}}$ is a learnable projection weight. 
In comparison with the attention-based sampler, it is simple and compact having a small number of parameters, but there is no data-adaptive process.
Due to its simplicity, similar approaches have been broadly used in the previous works such as SENet~\cite{SENET} and CBAM~\cite{CBAM}.

\textbf{Attention-based sampler }
The attention mechanism operates by mapping a query vector and a set of interchangeable key-value vector pairs into a single vector, which is a weighted sum of value vectors.
It allows us to aggregate a set of local descriptors by sparsely and adaptively selecting them. 
We hence apply this approach to our concept sampler.
For each concept $c$, the query vector $q_c \in \mathbb{R}^{1 \times \tilde{p}}$ describes what to focus on during aggregation. 
The concept-wise feature map $Z_c$ converts into a set of key-value vector pairs that we separately project it into a key map $K_c = Z_c W^{\text{k}}_c $ and a value map $V_c = Z_c  W^{\text{v}}_c$, where $W^{\text{k}}_c, W^{\text{v}}_c \in \mathbb{R}^{p \times \tilde{p}}$ are learnable projection weights.
The concept state $h_c$ is derived by computing the dot products of the query vector $q_c$ with the key map $K_c$ and subsequently applying a softmax function to obtain the attention weights over the value map $V_c$ as:
\begin{equation}\label{eqn:attn_cs}
h_c =F_{\text{CS}}(Z_c; \theta^{\text{CS}}_c) = \text{softmax} \left( \frac{q_c K_c^{\top}}{\sqrt{\tilde{p}}}\right)V_c = \left(\text{softmax} \left( \frac{q_c \left( Z_c W^{\text{k}}_c \right)^{\top}}{\sqrt{\tilde{p}}}\right)  Z_c\right) W^{\text{v}}_c.
\end{equation}
The query vector $q_c$ can be either learned as a model parameter or computed by a function of the feature map $Z_c$.
The former approach defines a static query that is shared globally over all data.
GCNet~\cite{GCNET} uses this approach, instead of global average pooling, to extract the global context.
It can be simplified and implemented by replacing the term $q_c \left( Z_c W^{\text{k}}_c \right)^{\top}$ in Equation~\ref{eqn:attn_cs} with a $1 \times 1$ convolution as depicted in Figure \ref{fig:cs}.b.
The latter approach, in contrast, uses a dynamic query that varies according to $Z_c$.
We set the query as an output of the function as $q_c = F_{\text{GAP}}(Z_c) W^{\text{q}}_c$, which is equal to the pool-based sampler, as shown in Figure \ref{fig:cs}.c.

The concept samplers can be viewed as multi-head attention modules in Transformer models~\cite{NIPS2017_7181} that we set each concept to be operated by a single-head attention module. 
However, our concept samplers don't process the same input feature map as they do.
Each concept is only accessible to its corresponding feature map, and this encourages the concept samplers to attend and process different features.

\subsection{Concept Reasoner}
\begin{figure}[t]
\makebox[\textwidth][c]{
\begin{minipage}[t]{.3\linewidth}
\centering
\includegraphics[width=0.8\textwidth]{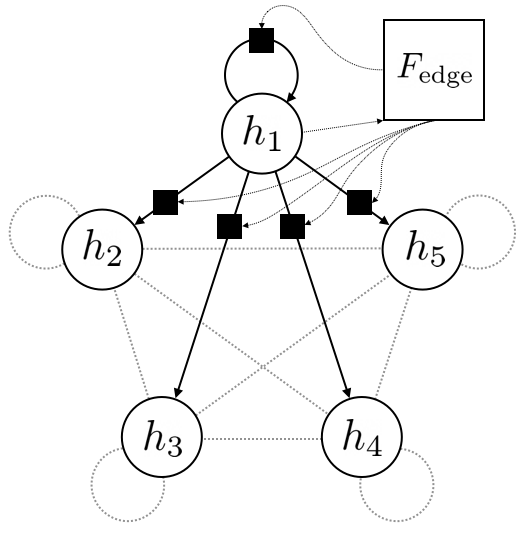}
\caption{Concept reasoner}
\label{fig:cr}
\end{minipage}
\hfill
\begin{minipage}[t]{.8\linewidth}
\centering
\includegraphics[width=0.9\textwidth]{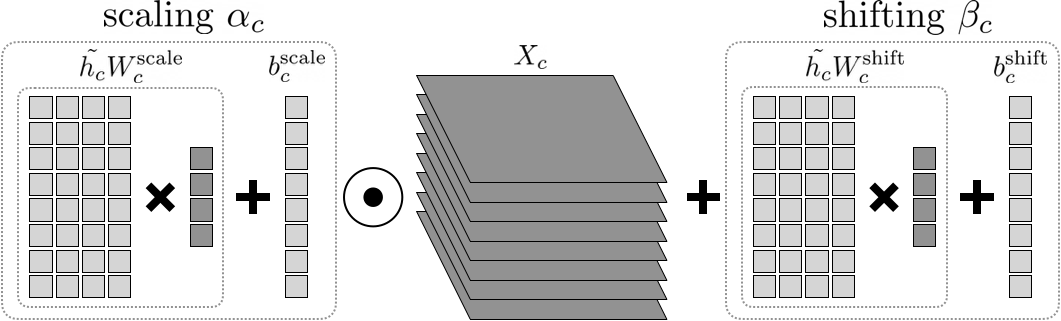}
\caption{Concept modulator}
\label{fig:cm}
\end{minipage}
}
\end{figure}

The visual concept states are derived independently from separate branches in which no communication exists.
Therefore, we introduce a reasoning module, \textit{Concept Reasoner} (CR), to make the visual concept states to interact with the others and accordingly update them.
We opt for using a graph-based method by defining a fully-connected graph $\mathcal{G} = (\mathcal{V}, \mathcal{E})$ with nodes $\mathcal{V}$ and directional edges $\mathcal{E}$.
The node $v_c \in \mathcal{V}$ corresponds to a single visual concept $c$ and is described by the visual concept state $h_c$.
The edge $e_{cc'} \in \mathcal{E}$ defines the relationship or dependency between visual concepts $c$ and $c'$.
It is further specified by an adjacency matrix $A \in \mathbb{R}^{C \times C}$ to represent edge weight values in a matrix form.
Based on this setting, we describe the update rule of the visual concept states as:
\begin{equation}\label{eqn:cr}
\tilde{h}_c = F_{\text{CR}}(h_c, \{h_{c'}\}_{c'=1}^{C};\theta_c^{\text{CR}})= \text{ReLU}\left(\text{BN}\left(h_c + \sum_{c'=1}^{C}A[c, c']{h_{c'}}\right)\right),    
\end{equation}
where $A[c, c'] \in \mathbb{R}$ is a edge weight value, and batch normalization (BN) and ReLU activation are used. 
This can also be implemented in a matrix form as $\tilde{H} = \text{ReLU}(\text{BN}(H + AH))$, where $H=[h_1;h_2;...;h_C] \in \mathbb{R}^{C \times \tilde{p}}$ is of vertically stacked concept states. 
The adjacency matrix $A$ can be treated as a module parameter that is learned during training.
This sets the edges to be static that all relationships between visual concepts are consistently applied to all data. 
However, we relax this constraint by dynamically computing the edge weights based on the concept states.
A function $F_{\text{edge}}(h_c;W^{\text{edge}})=\text{tanh}(h_c W^{\text{edge}})$, where $W^{\text{edge}} \in \mathbb{R}^{\tilde{p} \times C}$ is a learnable projection weight, is used to get all edge weights $A[c,:]$ related to the concept $c$ as shown in Figure \ref{fig:cr}. 
This function learns how each concept adaptively relates to the others based on its state.

\subsection{Concept Modulator}
The updated concept states are regarding not only a single concept, but also the others as a result of reasoning based on interactions. 
This information has to be further propagated to local concept features, which are extracted from the mainstream of the network. 
However, this is a non-trivial problem due to dimensional mismatch that the concept states are vectors not containing any explicit spatial information.
We alleviate this issue by implementing a module, \textit{Concept Modulator} (CM), which is based on a feature modulation approach.
It modulates the concept-wise feature maps by channel-wise scaling and shifting operations.
These operations are conditioned on the updated concept states to fine-tune the feature maps based on the result of reasoning.
We design this module based on DLN~\cite{DLN} and FiLM~\cite{FILM}.
Both models use feature-wise affine transformations on normalization layers by dynamically generating the affine parameters instead of learning them. 
In this way, we define separate modules for the visual concepts as shown in Figure~\ref{fig:cm}.
Each concept-wise feature map $X_c$ is modulated as:
\begin{gather}
    \tilde{X}_c = F_{\text{CM}}(\tilde{h}_c, X_c; \theta_c^{\text{CM}})= \text{ReLU}\left(\alpha_c \odot X_{c} + \beta_c\right), \\
    \alpha_c =\tilde{h_c}W^{\text{scale}}_c + b_{c}^{\text{scale}} \text{ and } \beta_c = \tilde{h_c}W^{\text{shift}}_c + b_{c}^{\text{shift}}, \label{eqn:cm_map}
\end{gather}
where $\odot$ indicates channel-wise multiplication.
$\alpha_c, \beta_c \in \mathbb{R}^{1 \times p}$ are scaling and shifting parameters, respectively, which are adaptively computed by linerly mapping the updated concept state $\tilde{h}_c$.

\section{Experiments}
In this section, we run experiments on visual recognition tasks such as image classification, object detection/segmentation, scene recognition, and action recognition with large-scale datasets. 
In all experiments, we set ResNeXt~\cite{resnext}, which performs better than ResNet~\cite{resnet} with less parameters, as a base architecture with cardinality $= 32$ and base width $=4$d.
Furthermore, our proposed model, VCRNet, is also defined by $C=32$ concepts in all residual blocks.
We also compare VCRNet against other networks (modules), which have a form of attention or reasoning modules, such as Squeeze-and-Excitation (SE)~\cite{SENET}, Convolutional Block Attention Module (CBAM)~\cite{CBAM}, Global Context block (GC)~\cite{GCNET}, and Global Reasoning unit (GloRe)~\cite{glore}.
All networks are implemented in all residual blocks in the ResNeXt except GloRe, which is partially adopted in the second and third residual stages.

\begin{table}[t]
\centering
\caption{Results of image classification on ImageNet validation set}
\resizebox{0.65\linewidth}{!}{
\begin{tabular}{l|c|c|c|c}
\hline
\multicolumn{1}{c|}{\multirow{2}{*}{Model}} & \multicolumn{2}{c|}{Error (\%)}                         & \multicolumn{1}{c|}{\multirow{2}{*}{\begin{tabular}[c]{@{}c@{}}\# of\\ Params\end{tabular}}} & \multicolumn{1}{c}{\multirow{2}{*}{GFLOPs}} \\ \cline{2-3}
\multicolumn{1}{c|}{}                       & \multicolumn{1}{c|}{Top-1} & \multicolumn{1}{c|}{Top-5} & \multicolumn{1}{c|}{}                                                                        & \multicolumn{1}{c}{}            
\\ \hline
ResNeXt-50~\cite{resnext} & 21.10 & 5.59 & 25.03M & 4.24 \\       
ResNeXt-50 + SE~\cite{SENET} & 20.79 & 5.38 & 27.56M & 4.25 \\
ResNeXt-50 + CBAM~\cite{CBAM} & 20.73 & 5.36 & 27.56M & 4.25 \\
ResNeXt-50 + GC~\cite{GCNET} & 20.44 & 5.34 & 27.58M & 4.25 \\
ResNeXt-50 + GloRe~\cite{glore} & 20.15 & 5.14 & 30.79M & 5.86 \\
ResNeXt-50 + VCR (ours) & 19.97 & 5.03 & 25.26M & 4.26 \\
ResNeXt-50 + VCR (pixel-level) & 19.94 & 5.18 & 25.26M & 4.29 \\
\hline
ResNeXt-101~\cite{resnext} & 19.82 & 4.96 & 44.18M & 7.99 \\
ResNeXt-101 + SE~\cite{SENET} & 19.39 & 4.73 & 48.96M & 8.00 \\
ResNeXt-101 + CBAM~\cite{CBAM} & 19.60 &  4.87 & 48.96M & 8.00  \\
ResNeXt-101 + GC~\cite{GCNET} & 19.52 & 5.03 & 48.99M & 8.00  \\
ResNeXt-101 + GloRe~\cite{glore} & 19.56 & 4.85 & 49.93M & 9.61  \\
ResNeXt-101 + VCR (ours) & 18.84 & 4.48 & 44.60M & 8.01 \\
\hline
\end{tabular}
}
\label{tab:imagenet}
\end{table}

\subsection{Image Classification}\label{sec:imagenet}
We conduct experiments on a large-scale image classification task on the ImageNet dataset~\cite{ILSVRC15}. 
The dataset consists of 1.28M training images and 50K validation images from 1000 different classes.
All networks are trained on the training set and evaluated on the validation set by reporting the top-1 and top-5 errors with single center-cropping.
Our training setting is explained in detail in Appendix.

The overall experimental results are shown in Table~\ref{tab:imagenet}, where all results are reproduced by our training setting for a fair comparison.
For evaluation, we always take the final model, which is obtained by exponential moving average (EMA) with the decay value 0.9999.
VCRNet consistently outperforms than other networks in both ResNeXt-50 and ResNeXt-101 settings.
Moreover, it is more compact than the others that only increases the number of parameters by less than $1\%(\approx 0.95\%)$.
In contrast, GloRe~\cite{glore}, which also does high-level reasoning as our model, requires more parameters than ours, although it is partially applied in the ResNeXt architecture. 
In addition, we modify the concept modulators to reuse the attention maps extracted from the concept samplers to modulate local descriptors at \textit{pixel-level} as GloRe has a pixel-level re-projection mechanism.
The modification slightly improves the top-1 performance by using the same number of parameters, but it increases the computational cost (GFLOPs).
We describe this approach in detail in Appendix.

\subsection{Object Detection and Segmentation}
We further do some experiments on object detection and instance segmentation on the MSCOCO 2017 dataset~\cite{COCO}. 
MSCOCO dataset contains 115K images over 80 categories for training, 5K for validation. 
Our experiments are based on the Detectron2~\cite{detectron2}.
All backbone networks are based on the ResNeXt-50 and pre-trained on the ImageNet dataset by default.
We employ and train the Mask R-CNN with FPN~\cite{MASKRCNN}.
We follow the training procedure of the Detectron2 and use the $1\times$ schedule setting. 
Furthermore, synchronized batch normalization is used instead of freezing all related parameters.

For evaluation, we use the standard setting of evaluating object detection and instance segmentation via the standard mean average-precision scores at different boxes and the mask IoUs, respectively. 
Table~\ref{tab:coco} is the list of results by only varying the backbone network.
It shows similar tendencies to the results of ImageNet.
However, GloRe~\cite{glore} is showing the best performance.
We assume that this result is from two factors.
One is the additional capacity, which is relatively larger than other models, used by Glore.
The other is that GloRe uses pixel-level re-projection mechanism that applies the result of reasoning by re-computing all local descriptors. 
Especially, the task requires to do prediction on pixel-level so that it would be beneficial to use it.
Therefore, we also make our model to use pixel-level feature modulation, and it further improves the performance.

\begin{table}[t]
\caption{Results of object detection and instance segmentation on COCO 2017 validation set}
\centering
\resizebox{1.0\linewidth}{!}{
\begin{tabular}{l|ccc|ccc|c}
\hline
\multicolumn{1}{c|}{Backbone Network} & \multicolumn{1}{c}{$\text{AP}^\text{bbox}$} & \multicolumn{1}{c}{$\text{AP}_{50}^\text{bbox}$} & \multicolumn{1}{c|}{$\text{AP}_{75}^\text{bbox}$} & \multicolumn{1}{c}{$\text{AP}^\text{mask}$} & \multicolumn{1}{c}{$\text{AP}_{50}^\text{mask}$} & \multicolumn{1}{c|}{$\text{AP}_{75}^\text{mask}$} & \multicolumn{1}{c}{\# Params} \\ \hline
ResNeXt-50~\cite{resnext}& 40.70 & 62.02 & 44.49 & 36.75 & 58.89 & 39.03 & 43.94M \\
ResNeXt-50 + SE~\cite{SENET}& 41.04 & 62.61 & 44.45 & 37.13 & 59.53 & 39.79 & 46.47M \\
ResNeXt-50 + CBAM~\cite{CBAM}& 41.69 & 63.54 & 45.17 & 37.48 & 60.27 & 39.71 & 46.47M \\
ResNeXt-50 + GC~\cite{GCNET}& 41.66 & 63.76 & 45.29 & 37.58 & 60.36 & 39.92 & 46.48M \\
ResNeXt-50 + GloRe~\cite{glore}& 42.31 & 64.18 & 46.13 & 37.83 & 60.63 & 40.17 & 49.71M \\
ResNeXt-50 + VCR (ours)& 41.81 & 63.93 & 45.67 & 37.71 & 60.36 & 40.25 & 44.18M \\ 
ResNeXt-50 + VCR (pixel-level)& 42.02 & 64.15 & 45.87 & 37.75 & 60.62 & 40.22 & 44.18M \\
\hline
\end{tabular}
}
\label{tab:coco}
\end{table}

\begin{table}[t]
\begin{minipage}[t]{.5\linewidth}
\captionsetup{justification=centering}
\caption{Results of scene recognition \\ on Places-365 validation set} \label{tab:scene}
\centering
\resizebox{\linewidth}{!}{
    \begin{tabular}{l|c|c|c}
    \hline
    \multicolumn{1}{c|}{\multirow{2}{*}{Model}} & \multicolumn{2}{c|}{Error (\%)}                         & \multicolumn{1}{c}{\multirow{2}{*}{\begin{tabular}[c]{@{}c@{}}\# of\\ Params\end{tabular}}} \\ \cline{2-3}
    \multicolumn{1}{c|}{} & \multicolumn{1}{l|}{Top-1} & \multicolumn{1}{l|}{Top-5} & \multicolumn{1}{c}{}\\ \hline
    ResNeXt-50~\cite{resnext}  & 43.49 & 13.54 & 23.73M \\
    ResNeXt-50 + SE~\cite{SENET}  & 43.18&  13.41  & 26.26M \\
    ResNeXt-50 + CBAM~\cite{CBAM}&  43.18&  13.45& 26.26M \\
    ResNeXt-50 + GC~\cite{GCNET}  & 43.07& 13.34& 26.28M \\
    ResNeXt-50 + GloRe~\cite{glore}  & 42.94 & 13.22 & 29.48M \\
    ResNeXt-50 + VCR (ours)  & 42.92 & 12.96 & 23.96M \\ \hline
    \end{tabular}  
}
\end{minipage}%
\hfill
\begin{minipage}[t]{.5\linewidth}
\captionsetup{justification=centering}
\caption{Results of action recognition \\
on Kinetics-400 validation set}  \label{tab:action}
\centering
\resizebox{0.99\linewidth}{!}{%
    \begin{tabular}{l|c|c|c}
    \hline
    \multicolumn{1}{c|}{\multirow{2}{*}{\begin{tabular}[c]{@{}c@{}}Backbone network\\ (Slow-only pathway)\end{tabular}}} & \multicolumn{2}{c|}{Error (\%)}                         & \multicolumn{1}{c}{\multirow{2}{*}{\begin{tabular}[c]{@{}c@{}}\# of\\ Params\end{tabular}}} \\ \cline{2-3}
    \multicolumn{1}{c|}{} & \multicolumn{1}{l|}{Top-1} & \multicolumn{1}{l|}{Top-5} & \multicolumn{1}{c}{}\\ \hline
    ResNeXt-50~\cite{resnext}  & 26.41 & 9.43 & 40.07M \\
    ResNeXt-50 + SE~\cite{SENET}  & 25.06 & 8.70 & 42.58M \\
    ResNeXt-50 + CBAM~\cite{CBAM}  & 24.87 & 8.81 & 42.59M \\
    ResNeXt-50 + GC~\cite{GCNET}  & 25.31 & 9.32 & 42.60M \\
    ResNeXt-50 + GloRe~\cite{glore}  & 25.52 & 9.23 & 45.81M \\
    ResNeXt-50 + VCR(ours) & 24.73 & 8.39 & 40.28M \\ \hline
    \end{tabular}   
}
\end{minipage}%
\end{table}

\subsection{Scene Recognition and Action Recognition}
Places365~\cite{places} is a dataset labeled with scene semantic categories for the scene recognition task. 
This task is challenging due to the ambiguity between classes that several scene classes may share some similar objects causing confusion among them.
We specifically use the Places365-Standard setting that the train set has up to 1.8M images from 365 scene classes, and the validation set has 50 images per each class.
All networks are trained from random initialization and evaluated on the validation set by following the setting used in our ImageNet experiments.
Additionally, we insert Dropout~\cite{dropout} layers in residual blocks with $p=0.02$ to avoid some over-fitting. 

The human action recognition task is another task appropriate to demonstrate how the network can generalize well not only to 2D image data, but also to 3D video data.
We use the Kinetics-400 dataset~\cite{KINETICS} including 400 human action categories with 235K training videos and 20K validation videos. 
We follow the slow-only experiment setting used in \cite{slowfast} that simply takes the ImageNet pre-trained model with a parameter inflating approach~\cite{inflat}.

Both tasks are classification tasks similar to the ImageNet image classification, and the results shown in Table~\ref{tab:scene} and \ref{tab:action} explain that our approach are generally performing better than other baselines in various visual classification tasks.
Moreover, action recognition results prove that our model can be generally applied to all types of data.

\subsection{Analysis: Ablation Study and Visualization}
\textbf{Concept sampler } 
We have proposed different approaches for the concept sampler in Section~\ref{sec:cs}. 
To compare these approaches, we train our proposed networks (ResNeXt-50 + VCR) by having different concept samplers and keeping all other modules fixed.
Table~\ref{tab:ablation}.(a) compares the performance of these approaches on the ImageNet image classification task. 
The attention-based approach with dynamic queries (dynamic attn) outperforms the others, and we assume that this is due to having more adaptive power than the others.
Furthermore, the results interestingly show that our models consistently perform better than other baseline networks except a network with GloRe, which are shown in Table~\ref{tab:imagenet}, regardless of the type of concept sampler.

\textbf{Concept reasoner } 
To investigate the effectiveness of reasoning based on interactions between concepts, we conduct some experiments by modifying the concept reasoner.
We first remove the concept interaction term in Equation~\ref{eqn:cr} and evaluate it to measure the effectiveness of reasoning. 
Moreover, we also compare the performance between learned static edges and computed dynamic edges.
In Table~\ref{tab:ablation}.(b), the results show that the reasoning module is beneficial in terms of the performance.
Notably, it also reveals that using dynamic edges can improve the reasoning and reduce the number of parameters.

\textbf{Concept modulator }
Our feature modulation consists of both channel-wise scaling and shifting operations. 
Previous works have shown to use only scaling (gating)~\cite{SENET, CBAM, GENET} or only shifting~\cite{GCNET}.
We compare different settings of the feature modulation as shown in Table~\ref{tab:ablation}.(c). 
Using only shifting performs better than using only scaling, and combining both operations can be recommended as the best option.

\begin{figure}[t]
\centering
\includegraphics[width=\textwidth]{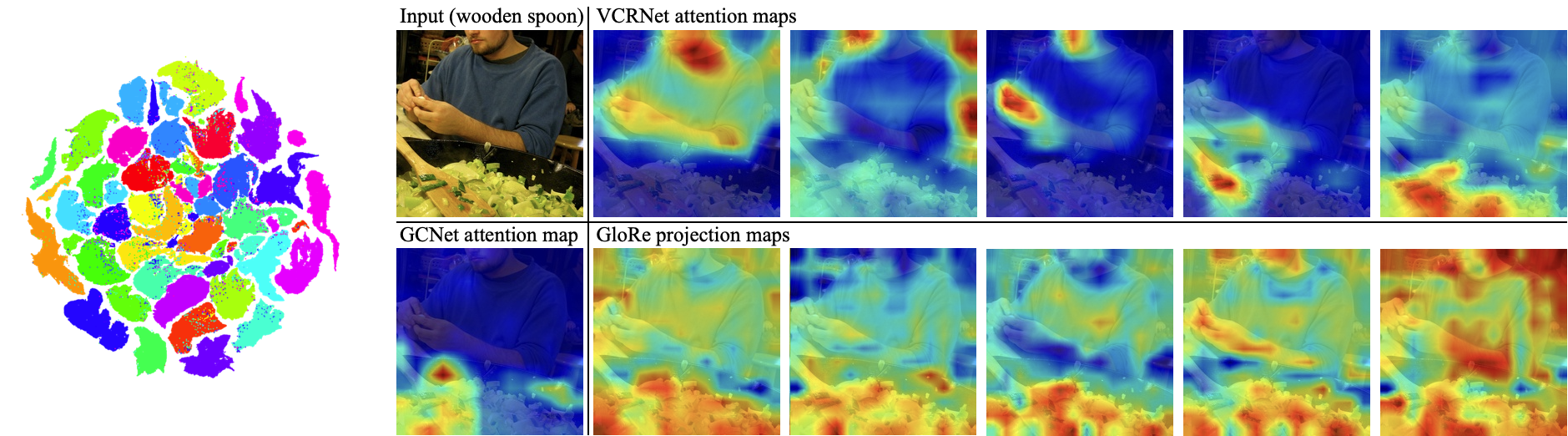}
\caption{(Left) t-SNE plots of visual concept states. $C=32$ concepts are distinguished by 32 colors. (Right) Visualization of attention (projection) maps from VCRNet, GCNet, and GloRe}
\label{fig:vis}
\vspace{-8pt}
\end{figure}

\textbf{Visualization}
We use t-SNE~\cite{tsne, tsne_gpu} to visual how visual concept states are existing in the feature space.
We collect a set of concept states, which are all extracted from the same concept sampler, by doing inference with the ImageNet validation set.
In Figure~\ref{fig:vis}, it is shown that the concept states are clustered and separated by concepts.
This result can be further explained by observing the attention maps computed from the concept samplers.
Interestingly, they reveal the fact that the concept samplers sparsely attend different regions or objects, and this would result in clustered concept states.
We also visualize attention (projection) maps from other networks such as GCNet~\cite{GCNET} and GloRe~\cite{glore} in Figure~\ref{fig:vis}.
GCNet only produces a single attention map, and it tends to sparsely attend foreground objects.
GloRe similarly computes projection maps as our approach, but the maps are densely attending regions with some redundancies between them.

\begin{table}[!t]
\captionsetup{skip=-5pt}
\caption{Ablation study on VCRNet}
\begin{minipage}[t]{.3333\linewidth}
\captionsetup{font=small,skip=1pt}
\caption*{(a) Concept Sampler}
\centering
 \resizebox{\linewidth}{!}{%

    \begin{tabular}{l|c|c}
    \hline
    \multicolumn{1}{c|}{\multirow{2}{*}{\begin{tabular}[c]{@{}c@{}}Model\\ (ResNeXt-50)\end{tabular}}} 
    & 
    \multicolumn{1}{c|}{Top-1}                         
    & 
    \multicolumn{1}{c}{\multirow{2}{*}{\begin{tabular}[c]{@{}c@{}}\# of\\ Params\end{tabular}}} 
    \\ 
    \multicolumn{1}{c|}{} 
    &
    \multicolumn{1}{c|}{Error (\%)}
    & 
    \multicolumn{1}{c}{}\\ 
    \hline
    pool  & 20.21 & 25.17M \\
    static attn  & 20.18 & 25.17M  \\
    dynamic attn & 19.97 & 25.26M \\ \hline
    \end{tabular}   
    
}%
\end{minipage}%
\hfill
\begin{minipage}[t]{.3333\linewidth}
\captionsetup{font=small,skip=1pt}
\caption*{(b) Concept Reasoner}
\centering
 \resizebox{\linewidth}{!}{%
        
    \begin{tabular}{l|c|c}
    \hline
    \multicolumn{1}{c|}{\multirow{2}{*}{\begin{tabular}[c]{@{}c@{}}Model\\ (ResNeXt-50)\end{tabular}}} 
    & 
    \multicolumn{1}{c|}{Top-1}                         
    & 
    \multicolumn{1}{c}{\multirow{2}{*}{\begin{tabular}[c]{@{}c@{}}\# of\\ Params\end{tabular}}} 
    \\ 
    \multicolumn{1}{c|}{} 
    &
    \multicolumn{1}{c|}{Error (\%)}
    & 
    \multicolumn{1}{c}{}\\ 
    \hline
    no edge  & 20.23 & 25.26M \\
    static edge  & 20.02 & 25.28M  \\
    dynamic edge & 19.97 & 25.26M \\ \hline
    \end{tabular}    
    
}%
\end{minipage}%
\hfill%
\begin{minipage}[t]{.3333\linewidth}
\captionsetup{font=small,skip=1pt}
\caption*{(c) Concept Modulator}
\centering
 \resizebox{\linewidth}{!}{%

        \begin{tabular}{l|c|c}
    \hline
    \multicolumn{1}{c|}{\multirow{2}{*}{\begin{tabular}[c]{@{}c@{}}Model\\ (ResNeXt-50)\end{tabular}}} 
    & 
    \multicolumn{1}{c|}{Top-1}                         
    & 
    \multicolumn{1}{c}{\multirow{2}{*}{\begin{tabular}[c]{@{}c@{}}\# of\\ Params\end{tabular}}} 
    \\ 
    \multicolumn{1}{c|}{} 
    &
    \multicolumn{1}{c|}{Error (\%)}
    & 
    \multicolumn{1}{c}{}\\ 
    \hline
    only scale       & 20.13 & 25.22M  \\
    only shift      & 20.05 & 25.22M \\ 
    scale + shift & 19.97 & 25.26M \\
    \hline
    \end{tabular}

  }%
\end{minipage}%
\label{tab:ablation}
\end{table}

\section{Conclusion}
In this work, we propose Visual Concept Reasoning Networks (VCRNet) that efficiently capture the global context by reasoning over high-level visual concepts.
Our proposed model precisely fits to a modularized multi-branch architecture by having split-transform-attend-interact-modulate-merge stages.
The experimental results shows that it consistently outperforms other baseline models on multiple visual recognition tasks and only increases the number of parameters by less than $1\%$.
We strongly believe research in these approaches will provide notable improvements on more difficult visual recognition tasks in the future.
As future works, we are looking forward to remove dense interactions between branches as possible to encourage more specialized concept-wise representation learning and improve the reasoning process.
Moreover, we expect to have consistent visual concepts that are shared and updated over all stages in the network.

\newpage


\bibliography{reference}

\begin{thebibliography}{10}\itemsep=-1pt

\bibitem{attention}
D.~Bahdanau, K.~Cho, and Y.~Bengio.
\newblock Neural machine translation by jointly learning to align and
  translate.
\newblock In Y.~Bengio and Y.~LeCun, editors, {\em 3rd International Conference
  on Learning Representations, {ICLR} 2015, San Diego, CA, USA, May 7-9, 2015,
  Conference Track Proceedings}, 2015.

\bibitem{AA}
I.~Bello, B.~Zoph, A.~Vaswani, J.~Shlens, and Q.~V. Le.
\newblock Attention augmented convolutional networks.
\newblock In {\em The IEEE International Conference on Computer Vision (ICCV)},
  October 2019.

\bibitem{Conscious}
Y.~Bengio.
\newblock The consciousness prior.
\newblock {\em CoRR}, abs/1709.08568, 2017.

\bibitem{GCNET}
Y.~{Cao}, J.~{Xu}, S.~{Lin}, F.~{Wei}, and H.~{Hu}.
\newblock Gcnet: Non-local networks meet squeeze-excitation networks and
  beyond.
\newblock In {\em 2019 IEEE/CVF International Conference on Computer Vision
  Workshop (ICCVW)}, pages 1971--1980, 2019.

\bibitem{inflat}
J.~{Carreira} and A.~{Zisserman}.
\newblock Quo vadis, action recognition? a new model and the kinetics dataset.
\newblock In {\em 2017 IEEE Conference on Computer Vision and Pattern
  Recognition (CVPR)}, pages 4724--4733, 2017.

\bibitem{tsne_gpu}
D.~M. Chan, R.~Rao, F.~Huang, and J.~F. Canny.
\newblock Gpu accelerated t-distributed stochastic neighbor embedding.
\newblock {\em Journal of Parallel and Distributed Computing}, 131:1--13, 2019.

\bibitem{glore}
Y.~Chen, M.~Rohrbach, Z.~Yan, Y.~Shuicheng, J.~Feng, and Y.~Kalantidis.
\newblock Graph-based global reasoning networks.
\newblock In {\em Proceedings of the IEEE Conference on Computer Vision and
  Pattern Recognition}, pages 433--442, 2019.

\bibitem{slowfast}
C.~{Feichtenhofer}, H.~{Fan}, J.~{Malik}, and K.~{He}.
\newblock Slowfast networks for video recognition.
\newblock In {\em 2019 IEEE/CVF International Conference on Computer Vision
  (ICCV)}, pages 6201--6210, 2019.

\bibitem{MASKRCNN}
K.~{He}, G.~{Gkioxari}, P.~{Dollár}, and R.~{Girshick}.
\newblock Mask r-cnn.
\newblock In {\em 2017 IEEE International Conference on Computer Vision
  (ICCV)}, 2017.

\bibitem{resnet}
K.~{He}, X.~{Zhang}, S.~{Ren}, and J.~{Sun}.
\newblock Deep residual learning for image recognition.
\newblock In {\em IEEE Conference on Computer Vision and Pattern Recognition
  (CVPR)}, 2016.

\bibitem{GENET}
J.~Hu, L.~Shen, S.~Albanie, G.~Sun, and A.~Vedaldi.
\newblock Gather-excite: Exploiting feature context in convolutional neural
  networks.
\newblock In S.~Bengio, H.~Wallach, H.~Larochelle, K.~Grauman, N.~Cesa-Bianchi,
  and R.~Garnett, editors, {\em Advances in Neural Information Processing
  Systems 31}, pages 9401--9411. Curran Associates, Inc., 2018.

\bibitem{SENET}
J.~{Hu}, L.~{Shen}, and G.~{Sun}.
\newblock Squeeze-and-excitation networks.
\newblock In {\em IEEE Conference on Computer Vision and Pattern Recognition
  (CVPR)}, 2018.

\bibitem{KINETICS}
W.~Kay, J.~Carreira, K.~Simonyan, B.~Zhang, C.~Hillier, S.~Vijayanarasimhan,
  F.~Viola, T.~Green, T.~Back, P.~Natsev, M.~Suleyman, and A.~Zisserman.
\newblock The kinetics human action video dataset.
\newblock {\em CoRR}, abs/1705.06950, 2017.

\bibitem{DLN}
T.~Kim, I.~Song, and Y.~Bengio.
\newblock Dynamic layer normalization for adaptive neural acoustic modeling in
  speech recognition.
\newblock In F.~Lacerda, editor, {\em Interspeech 2017, 18th Annual Conference
  of the International Speech Communication Association, Stockholm, Sweden,
  August 20-24, 2017}, pages 2411--2415. {ISCA}, 2017.

\bibitem{COCO}
T.-Y. Lin, M.~Maire, S.~Belongie, J.~Hays, P.~Perona, D.~Ramanan,
  P.~Doll{\'a}r, and C.~L. Zitnick.
\newblock Microsoft coco: Common objects in context.
\newblock In D.~Fleet, T.~Pajdla, B.~Schiele, and T.~Tuytelaars, editors, {\em
  European Conference Computer Vision {ECCV}}, 2014.

\bibitem{SGDR}
I.~Loshchilov and F.~Hutter.
\newblock {SGDR:} stochastic gradient descent with warm restarts.
\newblock In {\em International Conference on Learning Representations,
  {ICLR}}, 2017.

\bibitem{erf}
W.~Luo, Y.~Li, R.~Urtasun, and R.~Zemel.
\newblock Understanding the effective receptive field in deep convolutional
  neural networks.
\newblock In D.~D. Lee, M.~Sugiyama, U.~V. Luxburg, I.~Guyon, and R.~Garnett,
  editors, {\em Advances in Neural Information Processing Systems 29}, pages
  4898--4906. Curran Associates, Inc., 2016.

\bibitem{FILM}
E.~Perez, F.~Strub, H.~de~Vries, V.~Dumoulin, and A.~C. Courville.
\newblock Film: Visual reasoning with a general conditioning layer.
\newblock In {\em AAAI}, 2018.

\bibitem{ILSVRC15}
O.~Russakovsky, J.~Deng, H.~Su, J.~Krause, S.~Satheesh, S.~Ma, Z.~Huang,
  A.~Karpathy, A.~Khosla, M.~Bernstein, A.~C. Berg, and L.~Fei-Fei.
\newblock {ImageNet Large Scale Visual Recognition Challenge}.
\newblock {\em International Journal of Computer Vision (IJCV)}, 2015.

\bibitem{dropout}
N.~Srivastava, G.~Hinton, A.~Krizhevsky, I.~Sutskever, and R.~Salakhutdinov.
\newblock Dropout: A simple way to prevent neural networks from overfitting.
\newblock {\em Journal of Machine Learning Research}, 15(56):1929--1958, 2014.

\bibitem{inception}
C.~Szegedy, W.~Liu, Y.~Jia, P.~Sermanet, S.~Reed, D.~Anguelov, D.~Erhan,
  V.~Vanhoucke, and A.~Rabinovich.
\newblock Going deeper with convolutions.
\newblock In {\em Computer Vision and Pattern Recognition (CVPR)}, 2015.

\bibitem{LABELSMOOTH}
C.~Szegedy, V.~Vanhoucke, S.~Ioffe, J.~Shlens, and Z.~Wojna.
\newblock Rethinking the inception architecture for computer vision.
\newblock In {\em The IEEE Conference on Computer Vision and Pattern
  Recognition (CVPR)}, 2016.

\bibitem{tsne}
L.~van~der Maaten and G.~Hinton.
\newblock Visualizing data using {t-SNE}.
\newblock {\em Journal of Machine Learning Research}, 9:2579--2605, 2008.

\bibitem{NIPS2017_7181}
A.~Vaswani, N.~Shazeer, N.~Parmar, J.~Uszkoreit, L.~Jones, A.~N. Gomez, L.~u.
  Kaiser, and I.~Polosukhin.
\newblock Attention is all you need.
\newblock In I.~Guyon, U.~V. Luxburg, S.~Bengio, H.~Wallach, R.~Fergus,
  S.~Vishwanathan, and R.~Garnett, editors, {\em Advances in Neural Information
  Processing Systems 30}, pages 5998--6008. Curran Associates, Inc., 2017.

\bibitem{NLN}
X.~{Wang}, R.~{Girshick}, A.~{Gupta}, and K.~{He}.
\newblock Non-local neural networks.
\newblock In {\em 2018 IEEE/CVF Conference on Computer Vision and Pattern
  Recognition}, pages 7794--7803, 2018.

\bibitem{CBAM}
S.~Woo, J.~Park, J.~Lee, and I.~S. Kweon.
\newblock {CBAM:} convolutional block attention module.
\newblock In {\em European Conference Computer Vision {ECCV}}, 2018.

\bibitem{detectron2}
Y.~Wu, A.~Kirillov, F.~Massa, W.-Y. Lo, and R.~Girshick.
\newblock Detectron2.
\newblock \url{https://github.com/facebookresearch/detectron2}, 2019.

\bibitem{resnext}
S.~{Xie}, R.~{Girshick}, P.~{Dollár}, Z.~{Tu}, and K.~{He}.
\newblock Aggregated residual transformations for deep neural networks.
\newblock In {\em IEEE Conference on Computer Vision and Pattern Recognition
  (CVPR)}, 2017.

\bibitem{latentgnn}
S.~Zhang, X.~He, and S.~Yan.
\newblock {L}atent{GNN}: Learning efficient non-local relations for visual
  recognition.
\newblock In K.~Chaudhuri and R.~Salakhutdinov, editors, {\em Proceedings of
  the 36th International Conference on Machine Learning}, volume~97 of {\em
  Proceedings of Machine Learning Research}, pages 7374--7383, Long Beach,
  California, USA, 09--15 Jun 2019. PMLR.

\bibitem{cam}
B.~Zhou, A.~Khosla, L.~A., A.~Oliva, and A.~Torralba.
\newblock {Learning Deep Features for Discriminative Localization.}
\newblock {\em CVPR}, 2016.

\bibitem{places}
B.~Zhou, A.~Lapedriza, A.~Khosla, A.~Oliva, and A.~Torralba.
\newblock Places: A 10 million image database for scene recognition.
\newblock {\em IEEE Transactions on Pattern Analysis and Machine Intelligence},
  2017.

\end{thebibliography}
\bibliographystyle{neurips}

\newpage
\begin{appendices}
\section{Implementation Details}
\begin{figure}[h]
\centering
\includegraphics[width=\textwidth]{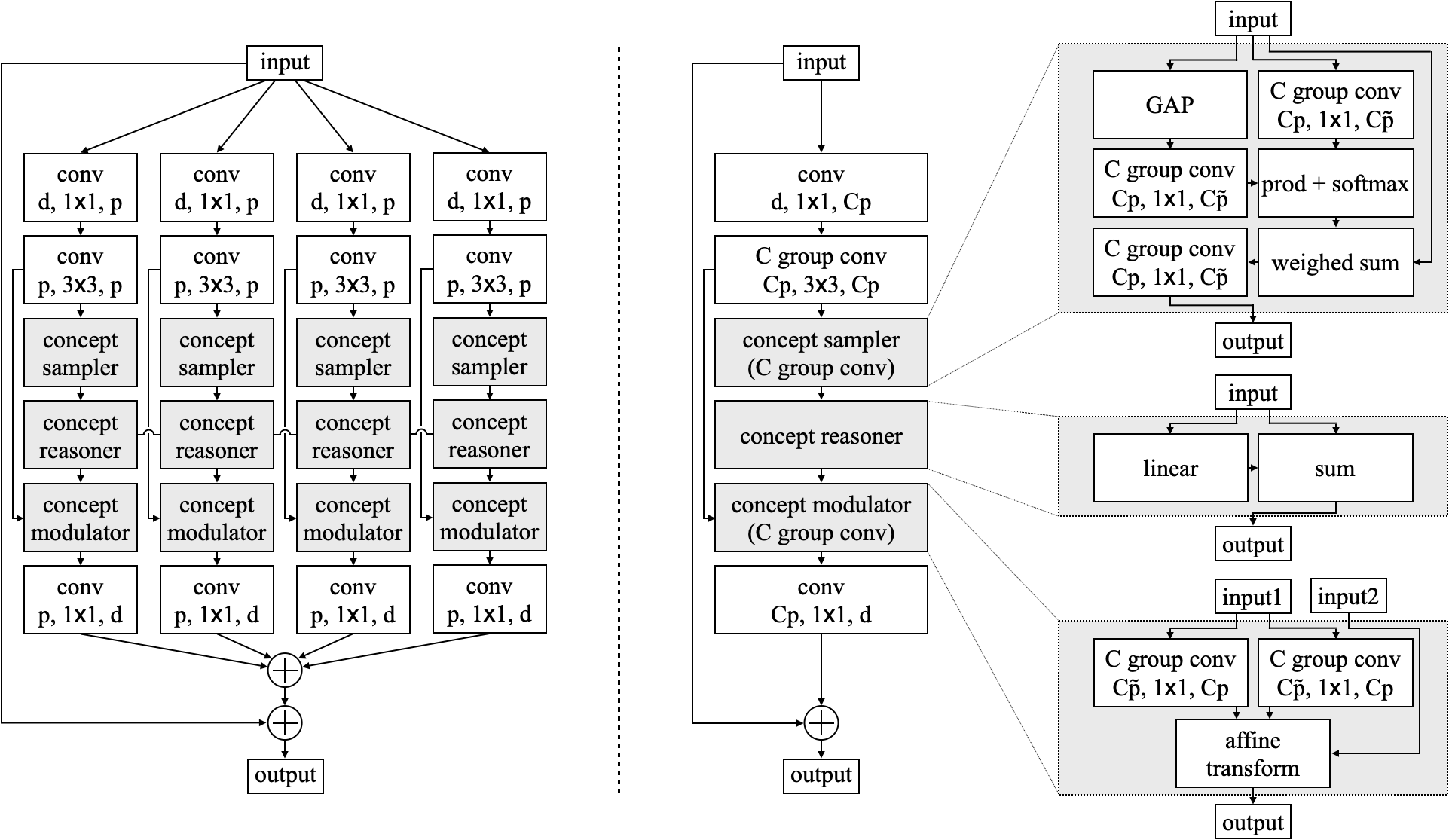}
\caption{Both architectures are equivalent. Left: Each branch is associated to a single concept $c$, and the concept reasoner is shared between branches. Right: By using grouped convolutions, the architecture can be implemented as a single-branch network. Reshaping operations are omitted in this figure. }
\label{fig:blk}
\end{figure}

Our proposed model is based on a modularized multi-branch architecture.
Each branch can be described as a separate network that is associated to a single concept as shown in Figure~\ref{fig:blk}.
The concept reasoner is only shared between branches to make different concepts to interact.
At the end, all branches are merged into a single output by summation.
By using grouped convolutional operations, the overall architecture can be viewed as a single branch, but internally operates by separate networks in parallel.

\section{Image Classification Training Setting}
For training the networks, we follow the standard data loader that uses data augmentation with resizing and random flipping and cropping to get images with the size of $224 \times 224$. 
All images are also normalized into $[0, 1]$ by using the RGB-wise mean and standard deviation. 
We train all networks from scratch on distributed learning system using synchronous SGD optimizer with weight decay 0.0001 and momentum 0.9. 
The learning rate is warmed up for the initial 5 epochs that is linearly increased from 0.0 to 1.6~\cite{SGDR}. 
This is the setting when the mini-batch size is 4096 with 32 GPUs (128 images per each GPU) for both ResNeXt-50 and ResNeXt-101. 
Afterward, we use the cosine learning rate scheduler for 115 epochs that decays the learning rate from 1.6 to 0.0001.
Moreover, we apply label-smoothing regularization~\cite{LABELSMOOTH} during training.

\section{Pixel-level Concept Modulator}
In this work, we mainly introduce a concept modulator with channel-level affine transformations that the related parameters are dynamically generated rather than learned.
This approach treats all local positions equally during feature modulation.
It is the simplest way to effectively propagate the global context into all local descriptors.
However, we further implement a pixel-level concept modulator to propagate the global context differently into local descriptors.
Each concept state $h_c$ is derived by computing the attention map $M_c \in \mathbb{R}^{HW \times 1}$ from the corresponding concept sampler that it contains the spatial information related to the concept $c$.
Therefore, we utilize this attention map for the pixel-level concept modulator.
We first re-normalize the attention map by its maximum value:
\begin{equation}
\tilde{M}_c = \frac{M_c}{\text{max}(M_c)} \text{, where } M_c = \text{softmax} \left( \frac{q_c \left( Z_c W^{\text{k}}_c \right)^{\top}}{\sqrt{\tilde{p}}}\right).    
\end{equation}
Without this re-normalization, the learning doesn't work properly.
The re-normalized attention map $\tilde{M}_c$ is used to project the updated concept state $\tilde{h}_c$ into all local positions as $\tilde{M}_c\tilde{h}_c \in \mathbb{R}^{HW \times \tilde{p}}$.
based on this projection, we are able to do pixel-level feature modulation as:
\begin{gather}
    \tilde{X}_c = F_{\text{CM}}(\tilde{h}_c, \tilde{M}_c, X_c; \theta_c^{\text{CM}})= \text{ReLU}\left(\alpha_c \cdot X_{c} + \beta_c\right), \\
    \alpha_c =\left(\tilde{M}_c\tilde{h}_c\right)W^{\text{scale}}_c + b_{c}^{\text{scale}} \text{ and } \beta_c = \left(\tilde{M}_c\tilde{h}_c\right)W^{\text{shift}}_c + b_{c}^{\text{shift}},
\end{gather}
where $\cdot$ is an element-wise multiplication.
Both $\alpha_c$ and $\beta_c$ are having the same size as the feature map $X_c$ so that all local positions have separate scaling and shifting parameters.

\section{Additional Experimental Results}
We do some additional experiments to investigate the effectiveness of normalization layers in our proposed modules.
In our main implementation, we use batch normalization (BN) layers in concept samplers and reasoners.
However, the effectiveness of batch normalization layers highly depends on the mini-batch size or the image size.
In image classification experiments, we set the mini-batch size significantly large, which is 4096, and the batch normalization layers seem to properly improve the performance as shown in Table~\ref{tab:imagenet_bn}.

\begin{table}[h]
\centering
\caption{The effectiveness of BN on image classification}
\begin{tabular}{l|c|c|c|c}
\hline
\multicolumn{1}{c|}{\multirow{2}{*}{\begin{tabular}[c]{@{}c@{}}Setting\\ (Model: ResNeXt-50 + VCR)\end{tabular}}} 
& \multicolumn{2}{c|}{Error (\%)}                         & \multicolumn{1}{c|}{\multirow{2}{*}{\begin{tabular}[c]{@{}c@{}}\# of\\ Params\end{tabular}}} & \multicolumn{1}{c}{\multirow{2}{*}{GFLOPs}} \\ \cline{2-3}
\multicolumn{1}{c|}{}                       & \multicolumn{1}{c|}{Top-1} & \multicolumn{1}{c|}{Top-5} & \multicolumn{1}{c|}{}                                                                        & \multicolumn{1}{c}{}            
\\ \hline
    remove BN (channel-level) & 20.15 & 5.12 & 25.25M & 4.26 \\
    remove BN (pixel-level) & 20.17 & 5.06 & 25.25M & 4.29 \\
    use BN (channel-level)  & 19.97 & 5.03 & 25.26M & 4.26 \\
    use BN (pixel-level)  & 19.94 & 5.18 & 25.26M & 4.29 \\
\hline
\end{tabular}
\label{tab:imagenet_bn}
\end{table}

We further run experiments on object detection and instance segmentation on the MSCOCO dataset.
All experiments are initialized by using the pre-trained models and trained with the mini-batch size 16 that is relatively smaller than the setting of image classification.
The results show that the batch normalization layers conversely degrade the performance in some cases.

\begin{table}[h]
\caption{The effectiveness of BN on object detection and instance segmentation}
\centering
\resizebox{1.0\linewidth}{!}{
\begin{tabular}{l|ccc|ccc|c}
\hline
\multicolumn{1}{c|}{Setting} & \multicolumn{1}{c}{$\text{AP}^\text{bbox}$} & \multicolumn{1}{c}{$\text{AP}_{50}^\text{bbox}$} & \multicolumn{1}{c|}{$\text{AP}_{75}^\text{bbox}$} & \multicolumn{1}{c}{$\text{AP}^\text{mask}$} & \multicolumn{1}{c}{$\text{AP}_{50}^\text{mask}$} & \multicolumn{1}{c|}{$\text{AP}_{75}^\text{mask}$} & \multicolumn{1}{c}{\# Params} \\ \hline
remove BN (channel-level) & 41.68 & 63.63 & 45.45 & 37.46 & 60.20 & 39.94 & 44.15M \\
remove BN (pixel-level) & 42.11 & 64.24 & 45.89 & 37.80 & 60.72 & 40.12 & 44.15M \\ 
use BN (channel-level) & 41.81 & 63.93 & 45.67 & 37.71 & 60.36 & 40.25 & 44.18M \\ 
use BN (pixel-level) & 42.02 & 64.15 & 45.87 & 37.75 & 60.62 & 40.22 & 44.18M \\
\hline
\end{tabular}
}
\label{tab:coco_bn}
\end{table}

\newpage
\section{Visualization: Class Activation Mapping}
\begin{figure}[h]
\centering
\includegraphics[width=\textwidth]{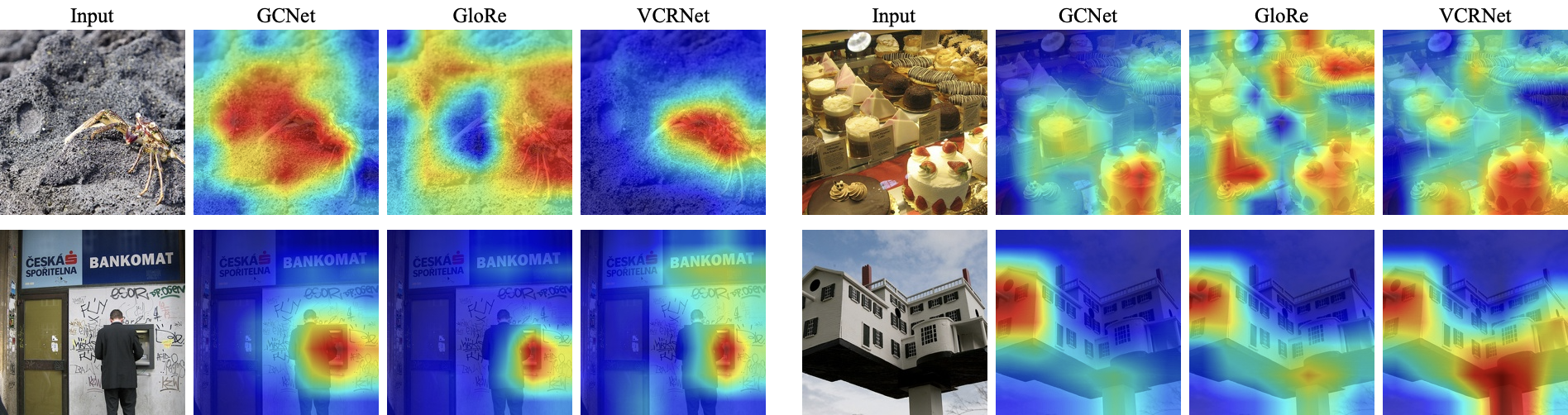}
\caption{Visualizations of class activation mapping (CAM) from different networks}
\label{fig:cam}
\end{figure}

We visualize class activation mapping (CAM)~\cite{cam} to localize class-specific regions in images.
The visualization results are shown in Figure~\ref{fig:cam}.

\end{appendices}

\end{document}